\documentclass[a4paper,twoside]{article}

\usepackage{epsfig}
\usepackage{subcaption}
\usepackage{calc}
\usepackage{amssymb}
\usepackage{amstext}
\usepackage{amsmath}
\usepackage{amsthm}
\usepackage{multicol}
\usepackage{pslatex}
\usepackage{apalike}
\usepackage{algorithm2e}
\usepackage[bottom]{footmisc}

\usepackage[pdftex, 
            pdfauthor={Vincent A. Cicirello},
            pdftitle={A Survey and Analysis of Evolutionary Operators for Permutations},
            pdfsubject={Cicirello, V.A. (2023). A Survey and Analysis of Evolutionary Operators for Permutations. In Proceedings of the 15th International Joint Conference on Computational Intelligence, pages 288-299. https://doi.org/10.5220/0012204900003595},
            pdfkeywords={Crossover, Evolutionary Algorithms, Fitness Landscape Analysis, Java, Mutation, Open Source, Permutation}]{hyperref}
\usepackage{fancyhdr}

\fancypagestyle{specialfooter}{%
  \fancyhf{}
  
  \fancyfoot[L]{Accepted version. Cite published version as:\\Cicirello, V.A. (2023). A Survey and Analysis of Evolutionary Operators for Permutations.  In {\it Proceedings of the 15th International Joint Conference on Computational Intelligence}, pages 288-299. \url{https://doi.org/10.5220/0012204900003595}}
  \fancyhead[L]{Preprint also available at: \url{https://www.cicirello.org/publications/cicirello2023ecta.pdf}}
}

\usepackage{SCITEPRESS}     

\begin{document}
\thispagestyle{specialfooter}

\title{A Survey and Analysis of Evolutionary Operators for Permutations}

\author{\authorname{Vincent A. Cicirello \orcidAuthor{0000-0003-1072-8559}}
\affiliation{Computer Science, School of Business, Stockton University, 101 Vera King Farris Dr, Galloway, NJ, USA}
\email{vincent.cicirello@stockton.edu}
}


\abstract{There are many combinatorial optimization problems whose solutions are
best represented by permutations. The classic traveling salesperson seeks an optimal 
ordering over a set of cities. Scheduling problems often seek optimal orderings of 
tasks or activities. Although some evolutionary approaches to such problems utilize 
the bit strings of a genetic algorithm, it is more common to directly represent 
solutions with permutations. Evolving permutations directly requires specialized 
evolutionary operators. Over the years, many crossover and mutation operators have 
been developed for solving permutation problems with evolutionary algorithms. In this 
paper, we survey the breadth of evolutionary operators for permutations. We implemented 
all of these in Chips-n-Salsa, an open source Java library for evolutionary 
computation. Finally, we empirically analyze the crossover operators on artificial 
fitness landscapes isolating different permutation features.}

\onecolumn \maketitle \normalsize \setcounter{footnote}{0} \vfill

\section{\uppercase{Introduction}}
\label{sec:introduction}

\noindent Combinatorial optimization often involves searching for an ordering of
a set of elements to either minimize a cost function or maximize a value function.
A classic example is the traveling salesperson (TSP), in which we seek the tour of 
a set of cities (i.e., simple cycle that includes all cities) that minimizes 
cost~\cite{Papadimitriou1998}. Many scheduling problems in a variety of domains 
also involve searching for an ordering of a set of elements (e.g., jobs, tasks, 
activities, etc) that either minimizes or maximizes some objective 
function~\cite{Huang2023,Ding2023,Xiong2022,Geurtsen2023,Li2022,Pasha2022}.

When faced with such an ordering problem, you can turn to evolutionary computation.
Although it is certainly possible to represent solutions to ordering problems with
bit strings to enable using a genetic algorithm and its standard operators, it is more 
common to more directly represent solutions with permutations. But to evolve a population
of permutations requires the use of specialized evolutionary operators.

Many crossover and mutation operators were developed over the years for evolving 
solutions to permutation problems. Some like Edge Recombination~\cite{Whitley1989} 
assume that a permutation represents edges such as for the TSP and are designed so 
children inherit edges from parents. Others like Precedence Preservative 
Crossover~\cite{Bierwirth1996} transfer precedences from parents to children, such 
as for scheduling problems where completing a given task earlier than others may 
improve solution fitness. Operators like Cycle Crossover~\cite{Oliver1987} enable 
children to inherit element positions from parents. Several studies examine the 
variety of problem features that can be important to solution fitness for permutation 
problems~\cite{Campos2005,cicirello2019bict,cicirello2022mone}. This is why the 
literature is so rich with crossover and mutation operators for permutations.

In this paper, we survey the breadth of evolutionary operators for permutations. We 
discuss how each operator works, runtime complexity, and the problem features that 
each operator focuses upon. Additionally, we implemented all of the 
operators in Chips-n-Salsa~\cite{cicirello2020joss}, an open source Java library for 
evolutionary computation, with source code available on GitHub 
(https://github.com/cicirello/Chips-n-Salsa). We specify our assumptions and other 
preliminaries in Section~\ref{sec:assumptions}. Sections~\ref{sec:mutation} and~\ref{sec:crossover} 
survey permutation mutation and crossover operators, respectively. In 
Section~\ref{sec:xoverlandscapes}, we perform fitness landscape analysis of the crossover 
operators on artificial landscapes that isolate different permutation features. This 
complements our prior analysis of mutation operators~\cite{cicirello2022mone}. We discuss
conclusions in Section~\ref{sec:conclusion}.

\section{\uppercase{Preliminaries}}
\label{sec:assumptions}

\noindent \textbf{Assumptions:} When discussing the runtime for implementations of an
operator, we assume that a permutation is represented as an array of elements.
The runtime of some operators may be different if permutations are instead
represented by linked lists. We assume both parents of a crossover are the 
same length, although some may generalize to different length parents. We 
use $n$ to denote permutation length.

We assume that operators alter the parent permutations, i.e., mutation 
mutates the input permutation and crossover transforms parents into 
children. In most cases, this does not impact algorithm complexity, but 
in other cases it does. The runtime of a few mutation operators is $O(1)$ 
with this assumption, but $O(n)$ if a new permutation was instead created.

\textbf{Permutation Features:} For each evolutionary operator, we discuss
the features of permutations that the operator enables offspring to inherit from
parents. We analyze the operational behavior of the crossover operators in this 
paper, including a fitness landscape analysis in Section~\ref{sec:xoverlandscapes}. 
For the mutation operators, some of the insights come from our prior work on 
fitness landscape analysis~\cite{cicirello2022mone}. The permutation features that we 
consider are as follows:
\begin{itemize}
\item Positions: Does mutation change a small number of element positions? Does crossover 
enable children to inherit absolute element positions from parents? For example, element 
$x$ is at index $i$ in a child if it was at index $i$ in one of its parents.
\item Undirected edges: If the permutation represents a sequence of undirected edges
(e.g., $x$ adjacent to $y$ implies an undirected edge between $x$ and $y$), does mutation
change a small number of edges? Does crossover enable inheriting undirected edges? For 
example, $x$ and $y$ are adjacent in a child if they are adjacent in one of the parents.
\item Directed edges: This is like above, but where a permutation
represents a sequence of directed edges, such as for the Asymmetric TSP (ATSP).
\item Precedences: Consider that a permutation represents a set of precedences. For example,
$x$ appearing anywhere prior to $y$ implies a preference for $x$ over $y$. Maybe a job for
a scheduling problem instance is critical to schedule earlier than others. Does mutation 
change a small number of precedences? Does crossover enable children to inherit precedences 
from parents?
\item Cyclic precedences: This feature is similar to the above, but such that the precedences
implied by the permutation follow some unspecified rotation.
\end{itemize}

\section{\uppercase{Mutation Operators}}
\label{sec:mutation}

\noindent There are many mutation operators for permutations. Some are so 
ubiquitous that it is difficult to attribute their origin to any specific 
work. Where possible, we cite the mutation operator's origin. Many of these,
however, are commonly utilized without specific attribution, and their origins 
likely lost to history. If additional reference is needed, there are many studies to 
consult~\cite{Bossek2023,cicirello2022mone,Sutton2014,Serpell2010,eiben03,Valenzuela2001}.

\textbf{Swap:} Swap mutation (also known as exchange) chooses two different
elements uniformly at random and swaps them. Its runtime is $O(1)$. It is a
good general purpose mutation because it is a small random change regardless
of the characteristics (e.g., positions, edges, precedences) most important to 
fitness.

\textbf{Adjacent Swap:} This is swap mutation restricted to adjacent elements. It 
tends to be associated with very slow search progress~\cite{cicirello2022mone} due
to a very small neighborhood size compared to other available mutation operators,
and is thus of more limited application. Its runtime is $O(1)$.

\textbf{Insertion:} Insertion mutation (also known as jump mutation) removes a
random element and reinserts it at a different randomly chosen position. Its worst 
case and average case runtime is $O(n)$ since all elements between the removal and 
insertion points shift ($n/3$ elements on average). It is worth considering in cases 
where the permutation represents a sequence of edges since an insertion is equivalent 
to replacing only 3 edges. It is shown especially effective when permutation element 
precedences are important to the problem~\cite{cicirello2022mone}. However, it is a poor 
choice when positions of elements impact fitness, because it disrupts the positions 
of a large number of elements ($n/3$ on average).

\textbf{Reversal and $2$-change:} Reversal mutation (also known as inversion) reverses 
a random sub-permutation. Its runtime is $O(n)$. Within 
a TSP context, reversal is approximately equivalent to $2$-change~\cite{Lin1965}, 
defined as replacing two edges of a tour to create a new tour. However, some reversals 
don't change any edges (e.g., reversing the entire permutation). We include both a 
reversal mutation and a true $2$-change mutation in the Chips-n-Salsa 
library~\cite{cicirello2020joss}. Reversal is a good choice when a permutation 
represents undirected edges. However, it is too disruptive in other cases, including 
when a permutation represents directed edges, such as for the ATSP, because it changes 
the direction of $n/3$ directed edges on average.

\textbf{3opt:} The 3opt mutation~\cite{Lin1965} was originally specified for the TSP
within the context of a steepest descent hill climber (i.e., systematically iterate over
the neighborhood). The 3opt neighborhood consists of all $2$-changes and $3$-changes,
where a $k$-change removes $k$ edges from a TSP tour and replaces them with $k$ edges that 
form a different valid tour. Our implementation in Chips-n-Salsa~\cite{cicirello2020joss} 
generalizes 3opt from TSP tours to permutations by assuming a permutation represents 
undirected edges, independent of what it actually represents, and then randomizes the 
``edge'' selection. Its runtime is $O(n)$. Like reversal, 3opt is worth considering 
when permutations represent undirected edges, but it is too disruptive in all other cases.

\textbf{Block-Move:} A block-move mutation removes a random contiguous block of
elements, and reinserts the block at a different random location. It generalizes
insertion mutation from single elements to a block of elements. Like insertion 
mutation, it is appropriate when permutations represent edge sequences (undirected or
directed) and it is equivalent to replacing three edges. It is too disruptive in other
contexts. Its worst case and average case runtime is $O(n)$.

\textbf{Block-Swap:} A block-swap (or block interchange) swaps two random
non-overlapping blocks. It generalizes swap mutation from elements to blocks. It also 
generalizes block-move, since a block-move swaps two adjacent blocks. Like block-move, 
it is appropriate when permutations represent edges (undirected or directed). In that 
context, it replaces up to four edges. It is too disruptive in other contexts. Its 
worst case and average case runtime is $O(n)$.

\textbf{Cycle:} Cycle mutation's two forms, $\mathit{Cycle}(\mathit{kmax})$ 
and $\mathit{Cycle}(\alpha)$, induce a random $k$-cycle~\cite{cicirello2022applsci}. 
They differ in how $k$ is chosen. $\mathit{Cycle}(\mathit{kmax})$ selects $k$ uniformly
at random from $\{2, \ldots, \mathit{kmax}\}$. $\mathit{Cycle}(\alpha)$ selects $k$
from $\{2, \ldots, n\}$, with probability of choosing $k=k^\prime$ proportional 
to $\alpha^{k^\prime-2}$ where $\alpha \in (0.0, 1.0)$. $\mathit{Cycle}(\alpha)$'s 
much larger neighborhood better enables local optima avoidance. The worst case and 
average runtime of $\mathit{Cycle}(\mathit{kmax})$ is a constant that depends 
upon $\mathit{kmax}$. The worst case runtime of $\mathit{Cycle}(\alpha)$ is $O(n)$, while 
average runtime is a constant that depends upon $\alpha$. Both forms are effective 
when element positions impact fitness. Fitness landscape analysis also suggests that it 
may be relevant in other cases provided $\mathit{kmax}$ or $\alpha$ carefully tuned.

\textbf{Scramble:} Scramble mutation, sometimes called shuffle, picks a random sub-permutation
and randomizes the order of its elements. Its worst case and average runtime is $O(n)$. It
is usually too disruptive. Surprisingly, it is effective for problems
where permutation element precedences are most important to fitness~\cite{cicirello2022mone},
likely because the vast majority of pair-wise precedences are retained (e.g., all involving at 
least one element not in the scrambled block, and on average half of the precedences where
both elements are in the scrambled block).

\textbf{Uniform Scramble:} We introduced uniform scramble mutation~\cite{cicirello2022mone} within
research on fitness landscape analysis. In the common form of scramble, the randomized elements 
are together in a block, while in uniform scramble they are distributed uniformly across the 
permutation. Each element is chosen with probability $u$, and the chosen elements are scrambled. 
Its worst case runtime is $O(n)$ and average case is $O(un)$. The work that introduced it 
considered the case of $u=1/3$ to affect the same number of elements on average as scramble. 
In that case, uniform scramble was too disruptive for general use. We posited that it may be useful to 
periodically kick a stagnated search (scramble might be useful for that purpose as well). Lower 
values of $u$ may be more effective for general use, but has not been explored.

\textbf{Rotation:} Chips-n-Salsa~\cite{cicirello2020joss} includes
a rotation mutation that performs a random circular rotation. It chooses the number of positions 
to rotate uniformly at random from $\{1, \ldots, n-1\}$. We have neither used it ourselves in 
research, nor have we seen others use it, so its strengths and weaknesses are unknown. It is 
not possible to transform one permutation into any other simply via a sequence of rotations. Thus, 
it is not likely effective as the only mutation operator in an EA, but might be useful in 
combination with others. Runtime is $O(n)$.

\textbf{Windowed Mutation:} Several mutation operators involve choosing two or three
random indexes into the permutation. Window-limited mutation constrains the distance between 
indexes~\cite{cicirello2014bict}. For example, window-limited swap
chooses two random elements that are at most $w$ positions apart. Windowed versions of swap, insertion, 
reversal, block-move, and scramble are all potentially applicable to some problems. Like adjacent swap,
which is a windowed swap with $w=1$, we posit that they likely lead to slow search progress in most cases
if $w$ is too low. We include them here to be comprehensive, but their strengths are unclear. The
runtime of windowed swap is $O(1)$, and the runtime of the others is $O(\min(n,w))$, or just
$O(w)$ if we assume that $w < n$.

\textbf{Algorithmic Complexity:} Table~\ref{tab:mutation} summarizes the worst case and 
average case runtime of the mutation operators. The bottom portion concerns the 
windowed mutation operators. The notation $\mathit{Swap}(w)$ means windowed 
swap with window limit $w$, and likewise for the other windowed operators.

\begin{table}[t]
\caption{Runtime of the mutation operators.}\label{tab:mutation}
\centering
\begin{tabular}{lll} \hline
Mutation & Worst case & Average case \\ \hline
Swap & $O(1)$ & $O(1)$ \\
Adjacent Swap & $O(1)$ & $O(1)$ \\
Insertion & $O(n)$ & $O(n)$ \\
Reversal & $O(n)$ & $O(n)$ \\
$2$-change & $O(n)$ & $O(n)$ \\
3opt & $O(n)$ & $O(n)$ \\
Block-Move & $O(n)$ & $O(n)$ \\
Block-Swap & $O(n)$ & $O(n)$ \\
$\mathit{Cycle}(\mathit{kmax})$ & $O((\frac{\mathit{kmax}}{2})^2)$ & $O((\frac{\mathit{kmax}}{4})^2)$ \\
$\mathit{Cycle}(\alpha)$ & $O(n)$ & $O((\frac{2-\alpha}{1-\alpha})^2)$ \\
Scramble & $O(n)$ & $O(n)$ \\
Uniform Scramble & $O(n)$ & $O(un)$ \\
Rotation & $O(n)$ & $O(n)$ \\ \hline
$\mathit{Swap}(w)$ & $O(1)$ & $O(1)$ \\
$\mathit{Insertion}(w)$ & $O(w)$ & $O(w)$ \\
$\mathit{Reversal}(w)$ & $O(w)$ & $O(w)$ \\
$\mathit{BlockMove}(w)$ & $O(w)$ & $O(w)$ \\
$\mathit{Scramble}(w)$ & $O(w)$ & $O(w)$ \\
\hline
\end{tabular}
\end{table}

\section{\uppercase{Crossover Operators}}
\label{sec:crossover}

\noindent There are many crossover operators for permutations. We describe the 
behavior of each here, and discuss the permutation features that they best enable 
inheriting from parents, confirmed empirically in Section~\ref{sec:xoverlandscapes}.

\textbf{Cycle Crossover (CX):} The CX~\cite{Oliver1987} operator picks a random 
permutation cycle of the pair of parents. Child $c_1$ inherits the positions of 
the elements that are in the cycle from parent $p_2$, and the positions of the 
other elements from parent $p_1$. Likewise, child $c_2$ inherits the positions 
of the elements in the cycle from $p_1$, and the positions of the other elements 
from $p_2$.

A permutation cycle is a cycle in a directed graph defined by a pair of permutations. 
The vertexes of this hypothetical graph are the permutation elements. This graph has 
an edge from $x$ to $y$ if and only if the index of $x$ in permutation $p_1$ is the 
same as the index of $y$ in permutation $p_2$. As an example, consider permutations 
$p_1 = [0, 1, 2, 3, 4, 5]$ and $p_2 = [2, 1, 4, 5, 0, 3]$. The graph consists of the 
directed edges: $\{ (0,2), (1, 1), (2,4), (3,5), (4,0), (5,3)\}$. We don't actually 
need to build the graph. There are three permutation cycles in this example: a $2$-cycle 
involving elements $\{3, 5\}$, a $3$-cycle involving elements $\{ 0, 2, 4\}$, and a 
singleton cycle with $\{1\}$.

CX picks a random index, computes the cycle containing the elements at that index, and 
exchanges those elements to form the children. In this example, if any of $\{ 0, 2, 4\}$ 
is the random element, then the children will be: $c_1 = [2, 1, 4, 3, 0, 5]$ and 
$c_2 = [0, 1, 2, 5, 4, 3]$. Runtime is $O(n)$.

CX strongly transfers positions from parents to children. Every element in a child inherits 
its position from one of the parents. Children also inherit all pairwise precedences 
from one or the other of the parents. For example, all elements in a child that inherited 
positions from $p_1$ retain the pairwise precedences from $p_1$, and similarly for the elements 
inherited from $p_2$. The cycle exchange likely breaks many edges.

\textbf{Edge Recombination (ER):} ER~\cite{Whitley1989} assumes that a permutation represents a 
cyclic sequence of undirected edges, such as for the TSP. Using a data structure called
an edge map~\cite{Whitley1989} for efficient implementation, the set of undirected edges that
appear in one or both parents is formed. Each child is then created only using edges from
that set. It is thus an excellent choice for problems where permutations represent undirected edges,
but it is unlikely to perform well otherwise. 

Consider an example with $p_1 = [3, 0, 2, 1, 4]$ and $p_2 = [4, 3, 2, 1, 0]$. Parent $p_1$ includes 
the undirected edges $\{ (3, 0), (0, 2), (2, 1), (1, 4), (4, 3) \}$, and $p_2$ includes 
$\{ (4, 3), (3, 2), (2, 1), (1, 0), (0, 4) \}$. The union of these is the set of undirected edges
$\{ (3, 0), (0, 2), (2, 1), (1, 4), (4, 3), (3, 2), (1, 0), (0, 4) \}$.
Initialize child $c_1$ with the first element of $p_1$ as $c_1 = [3]$. The 3 is adjacent to 0, 2, and 4
in the edge set. Choose the one that is adjacent to the fewest elements that are not yet in the child. In this
case, 0 is adjacent to three elements, and 2 and 4 are adjacent to two elements each. Break the tie randomly.
Consider that the random tie breaker resulted in 4 to give us $c_1 = [3, 4]$. The 4 is adjacent to 0 and 1 in 
the edge set, both of which are adjacent to two elements that haven't been used yet, so pick one of these
randomly. Assume that we chose 1 to arrive at $c_1 = [3, 4, 1]$. The 1 is adjacent to 0 and 2,
which are also the only remaining elements and thus another tie. Pick one at random, such as 0 in this example,
to obtain $c_1 = [3, 4, 1, 0]$. Finally add the last element: $c_1 = [3, 4, 1, 0, 2]$. The other child
$c_2$ is formed in a similar way, but initialized with the first element of $p_2$.
The runtime of ER is $O(n)$.

\textbf{Enhanced Edge Recombination (EER):} EER~\cite{Starkweather1991} works much like ER, forming children
from edges inherited from the parents. However, EER attempts to create children that inherit subsequences of 
edges that the parents have in common. Efficient implementation utilizes a variation of ER's edge map that 
is augmented to label the edges that the parents have in common. Each child is generated similarly to ER, 
except that the decision on which element to add next prefers edges that start common subsequences. See the 
original presentation of EER~\cite{Starkweather1991} or our open source implementation for full details. Like 
ER, the runtime of EER is $O(n)$ and it strongly enables inheriting undirected edges from parents, but not so 
much other permutation features.

\textbf{Order Crossover (OX):} OX~\cite{Davis1985} begins by selecting two random indexes to define a cross 
region similar to a two-point crossover for bit strings. Child $c_1$ gets the positions of the elements in 
the region from parent $p_1$, and the relative ordering of the remaining elements from $p_2$ but populated 
into $c_1$ beginning after the cross region in a cyclic manner. The original motivating problem was the TSP, 
which is likely why they chose to insert the relatively ordered elements after the cross region wrapping to 
the front since it gets all the relatively ordered elements together if you view the permutation as a cycle. 
Child $c_2$ is formed likewise, inheriting positions of elements in the region from $p_2$ and the relative 
order of the other elements from $p_1$. Runtime is $O(n)$.

Consider an example with $p_1 = [0, 1, 2, 3, 4, 5, 6, 7]$ and $p_2 = [1, 2, 0, 5, 6, 7, 4, 3]$. Let the 
random cross region consist of indexes 2 through 4. Child $c_1$ gets the elements at those indexes from $p_1$,
such as $c_1 = [x, x, 2, 3, 4, x, x, x]$, where each $x$ is a placeholder. The rest of the elements are 
relatively ordered as in $p_2$, i.e., in the order $1, 0, 5, 6, 7$, but populated into $c_1$ beginning after
the cross region to obtain $c_1 = [6, 7, 2, 3, 4, 1, 0, 5]$. Likewise, initialize $c_2$ with the elements
from the cross region of $p_2$ to get $c_2 = [x, x, 0, 5, 6, x, x, x]$. Then, populate it after the cross 
region with the remaining elements relatively ordered as in $p_1$ to obtain $c_2 = [4, 7, 0, 5, 6, 1, 2, 3]$.

OX is effective for edges, since it enables inheriting large numbers of edges. All adjacent elements 
within the cross region represent edges inherited from one parent. Getting the relative order of 
the remaining elements from the other parent tends to inherit many edges from that parent, although 
some edges will be broken where the cross region elements had been previously. Unlike ER and EER, OX's 
behavior for edges is independent of whether they are undirected or directed edges, such as for the TSP or 
the ATSP. OX is not effective for other features. Many precedences are flipped due to how OX populates 
the relatively ordered elements after the cross region wrapping to the front. And although the positions 
of the elements in the cross region are inherited by the children, the majority of the elements are 
positioned in the children rather differently than the parents. 

\textbf{Non-Wrapping Order Crossover (NWOX):} The NWOX operator~\cite{cicirello2006gecco} is similar
to OX. However, the relatively ordered elements populate the child from the left end of the permutation,
jumping over the cross region, and continuing to the right end. Consider the earlier example with parents
$p_1 = [0, 1, 2, 3, 4, 5, 6, 7]$ and $p_2 = [1, 2, 0, 5, 6, 7, 4, 3]$, and the same random cross region.
Child $c_1$ is still initialized with the cross region from $p_1$ as $c_1 = [x, x, 2, 3, 4, x, x, x]$. But
the relatively ordered elements, $1, 0, 5, 6, 7$, from $p_2$, fill in from the left to obtain
$c_1 = [1, 0, 2, 3, 4, 5, 6, 7]$. Likewise initialize $c_2 = [x, x, 0, 5, 6, x, x, x]$, but fill in
relatively ordered elements from the left to get $c_2 = [1, 2, 0, 5, 6, 3, 4, 7]$.

Unlike OX, NWOX is very effective for cases where element precedences are important to fitness, 
the original motivation for NWOX. Every pairwise precedence relation in a child is present in 
at least one of the parents. However, NWOX breaks more edges than OX. And like OX, NWOX tends to
displace elements from their original positions, although less so than OX. Runtime is $O(n)$.

\textbf{Uniform Order Based Crossover (UOBX):} UOBX~\cite{Syswerda1991} is a uniform analog of OX and NWOX.
It is controlled by a parameter $u$, which is the probability that a position is a fixed-point. That is,
child $c_i$ gets the position of the element at index $j$ in parent $p_i$ with probability $u$. The relative
order of the remaining elements comes from the other parent. Consider $p_1 = [3, 0, 6, 2, 5, 1, 4, 7]$ 
and $p_2 = [7, 6, 5, 4, 3, 2, 1, 0]$, $u=0.5$, and that indexes 0, 3, 4, and 6 were chosen
as fixed points. Child $c_1$ is initialized with $c_1 = [3, x, x, 2, 5, x, 4, x]$. The remaining elements 
are relatively ordered as in $p_2$ (i.e., 7, 6, 1, 0), and inserted into the open spots left to right to
obtain $c_1 = [3, 7, 6, 2, 5, 1, 4, 0]$. Likewise, $c_2$ gets the elements at the fixed points from
$p_2$: $c_2 = [7, x, x, 4, 3, x, 1, x]$. The other elements are then relatively ordered as in $p_1$ 
(i.e., 0, 6, 2, 5) to obtain $c_2 = [7, 0, 6, 4, 3, 2, 1, 5]$.

With UOBX, all pairwise precedences within the children are inherited from one or the other of the parents
(just like with NWOX). This is because the fixed points have same relative order as the parent where they 
originated, and the others the same relative order as in the other parent. For other problem features,
UOBX may be relevant if $u$ is carefully tuned. For example, UOBX is capable of transferring many element 
positions to children if $u$ is sufficiently high. UOBX likely breaks many edges since the fixed points 
are uniformly distributed along the permutation. However, if $u$ is low, it may retain many edges from 
parents. Runtime is $O(n)$.

\textbf{Order Crossover 2 (OX2):} Syswerda introduced OX2 in the same paper as UOBX~\cite{Syswerda1991}.
He originally called it order crossover, but others began using the name OX2~\cite{Starkweather1991} to 
distinguish it from the original OX, and that name has stuck. OX2 is rather different than OX. OX2 begins 
by selecting a random set of indexes. Syswerda's original description implied each index equally likely 
chosen as not chosen. In our implementation in Chips-n-Salsa~\cite{cicirello2020joss}, we 
provide a parameter $u$ that is the probability of choosing an index. For Syswerda's original OX2, set 
$u=0.5$. The elements at those indexes in $p_2$ are found in $p_1$. Child $c_1$ is formed as a copy of 
$p_1$ with those elements rearranged into the relative order from $p_2$. In a similar way, the elements 
at the chosen indexes in $p_1$ are found in $p_2$. Child $c_2$ becomes a copy of $p_2$ but with those 
elements rearranged into the relative order from $p_1$. Consider an example with 
$p_1 = [1, 0, 3, 2, 5, 4, 7, 6]$ and $p_2 = [6, 7, 4, 5, 2, 3, 0, 1]$, and random indexes: 1, 2, 6, and 7. 
The elements at those indexes in $p_2$, ordered as in $p_2$, are: 7, 4, 0, 1. Rearrange these within 
$p_1$ to produce $c_1 = [7, 4, 3, 2, 5, 0, 1, 6]$. The elements at the random indexes in $p_1$, ordered 
as in $p_1$, are: 0, 3, 7, 6. Rearrange these within $p_2$ to produce $c_2 = [0, 3, 4, 5, 2, 7, 6, 1]$. 

OX2 is closely related to UOBX. Each child produced by OX2 can be produced by UOBX from the same pair 
of parents, but different random indexes, and vice versa. However, the pair of children produced 
by OX2 will differ from the pair produced by UOBX. Thus, OX2 and UOBX are not exactly equivalent. 
It is unclear whether there are cases when one outperforms the other. They should be effective for 
the same problem features. OX2's runtime is $O(n)$.

\textbf{Precedence Preservative Crossover (PPX):} Bierwirth et al introduced two variations of a
crossover operator focused on precedences, both of which they referred to as 
PPX~\cite{Bierwirth1996}. We refer to the two-point version as PPX, and we consider the uniform
version later. Both variations were originally described as producing one child from two parents. 
In our implementation in Chips-n-Salsa~\cite{cicirello2020joss}, we generalize 
this to produce two children. In the two-point PPX, two random cross points, $i$ and $j$, are 
chosen. Without loss of generality, assume that $i$ is the lower of these. Child $c_1$ inherits 
everything left of $i$ from $p_1$, and likewise $c_2$ everything left of $i$ from $p_2$. Let 
$k=j-i+1$. Child $c_1$ inherits the first $k$ elements left-to-right from $p_2$ that are not 
yet in $c_1$, and similarly for the other child. The remaining elements of $c_1$ come from 
$p_1$ in the order they appear in $p_1$, and likewise for the other child.

Consider $p_1 = [7, 6, 5, 4, 3, 2, 1, 0]$ and $p_2 = [0, 1, 2, 3, 4, 5, 6, 7]$ with random $i=3$, 
$j=5$, and $k=j-i+1=3$. The first $i=3$ elements of $c_1$ come from $p_1$, i.e., $c_1 = [7, 6, 5]$, and
likewise $c_2 = [0, 1, 2]$. The next $k=3$ elements of $c_1$ are the first 3 elements of $p_2$ that are 
not yet in $c_1$, which results in $c_1 = [7, 6, 5, 0, 1, 2]$. Similarly, $c_2 = [0, 1, 2, 7, 6, 5]$. 
Complete $c_1$ with the remaining elements of $p_1$ that are not yet in $c_1$ from left to right. The 
final $c_1 = [7, 6, 5, 0, 1, 2, 4, 3]$ and $c_2 = [0, 1, 2, 7, 6, 5, 3, 4]$.

The children created by PPX inherit all precedences from one or the other of the parents. 
However, many edges are broken relative to the parents, and very few positions are inherited.
Runtime is $O(n)$.

\textbf{Uniform Precedence Preservative Crossover (UPPX):} We refer to the uniform version of 
PPX~\cite{Bierwirth1996} as UPPX. It originally produced one child from two parents, but we 
generalize to create two children. Generate a random array of $n$ booleans (Bierwirth et al 
originally specified array of ones and twos). Let $u$ be the probability of true. We added 
this $u$ parameter, which was not present in Bierwirth et al's version. For original UPPX, 
set $u=0.5$. To form child $c_1$ iterate over the array of booleans. If the next value is 
true, then add to $c_1$ the first element of $p_1$ not yet present in $c_1$, and otherwise 
add to $c_1$ the first element of $p_2$ not yet present. Form child $c_2$ at the same time, 
but true means add to $c_2$ the first element of $p_2$ not yet present in $c_2$, and 
otherwise the first element of $p_1$ not yet present.

UPPX focuses on transferring precedences from parents to children. The original case, $u=0.5$,
is unlikely suitable when edges or positions are important to fitness. However, higher or 
lower values of $u$ increases likelihood of consecutive elements coming from the same parent. 
Thus, tuning $u$ may lead to an operator relevant for problems like the TSP where edges are 
critical to fitness. Runtime is $O(n)$.

\textbf{Partially Matched Crossover (PMX):} Goldberg and Lingle introduced 
PMX~\cite{Goldberg1985,Goldberg1989}. PMX initializes children $c_1$ and $c_2$ 
as copies of parents $p_1$ and $p_2$. It defines a cross region with a random 
pair of indexes, which defines a sequence of swaps. For each index $i$ in the 
cross region, let $x$ be the element at index $i$ in $p_1$, and $y$ be the element at 
that position in $p_2$. Swap elements $x$ and $y$ within $c_1$, and swap $x$ and $y$ 
within $c_2$.

Consider $p_1 = [0, 1, 2, 3, 4, 5, 6, 7]$ and $p_2 = [1, 2, 0, 5, 6, 7, 4, 3]$,
and random cross region from index 2 to index 4, inclusive. Child $c_1$ is initially a 
copy of $p_1$: $c_1 = [0, 1, 2, 3, 4, 5, 6, 7]$. Swap the 2 with the 0 (i.e., elements at 
index 2 in the parents): $c_1 = [2, 1, 0, 3, 4, 5, 6, 7]$. Next, swap the 3 with the 5 (i.e., 
elements at index 3 in the parents): $c_1 = [2, 1, 0, 5, 4, 3, 6, 7]$. Finally, swap the 
4 with the 6 (i.e., elements at index 4 in the parents): $c_1 = [2, 1, 0, 5, 6, 3, 4, 7]$. 
Follow the same process for the other child to get: $c_2 = [1, 0, 2, 3, 4, 7, 6, 5]$.

In the original PMX description, the indexes of the elements to swap were found with a 
linear search, and since the average size of the cross region is also linear (i.e., 
$n/3$ on average), PMX as originally described required $O(n^2)$ time. The runtime of 
our implementation in Chips-n-Salsa~\cite{cicirello2020joss} is $O(n)$. 
Instead of linear searches, we generate the inverse of each permutation in $O(n)$ time 
to use as a lookup table to find each required index in constant time.

The cross region has an average of $n/3$ elements, due to which each child inherits 
$n/3$ element positions on average from each parent. Thus, children inherit a 
significant number of positions from the parents (at least $2n/3$ on average). 
Although the cross region of each child includes consecutive elements (i.e., edges) 
from the opposite parent, many edges are broken elsewhere, so PMX is disruptive for
edges. However, PMX preserves precedences well.

\textbf{Uniform Partially Matched Crossover (UPMX):} UPMX uses indexes uniformly 
distributed along the permutations rather than PMX's contiguous cross 
region~\cite{cicirello2000gecco}. Parameter $u$ is the probability of including an 
index. The elements at the chosen indexes define the swaps in the same way as in PMX. 
Runtime is $O(n)$.

Consider $p_1 = [7, 6, 5, 4, 3, 2, 1, 0]$ and $p_2 = [1, 2, 0, 5, 6, 4, 7, 3]$, and 
random indexes 3, 1, and 6. Initialize $c_1 = [7, 6, 5, 4, 3, 2, 1, 0]$. 
Element 4 is at index 3 in $p_1$, and 5 is at that index in $p_2$. UPMX swaps the 4 and 5
to obtain $c_1 = [7, 6, 4, 5, 3, 2, 1, 0]$. Elements 6 and 2 are at index 1 in $p_1$ and $p_2$.
UMPX swaps the 6 and 2 to get $c_1 = [7, 2, 4, 5, 3, 6, 1, 0]$. Elements 1 and 7 
are at index 6 in $p_1$ and $p_2$. Swap the 1 and 7 to get the final 
$c_1 = [1, 2, 4, 5, 3, 6, 7, 0]$. Follow the same process to derive  
$c_2 = [7, 6, 0, 4, 2, 5, 1, 3]$.

Children inherit positions and precedences from parents to about the same 
degree as in PMX. The uniformly distributed indexes likely break most edges. 
But if $u$ is low, many edges may also be preserved.

\textbf{Position Based Crossover (PBX):} Barecke and Detyniecki designed PBX
to strongly focus on positions~\cite{Barecke2007}. One of PBX's objectives is 
for a child to inherit approximately equal numbers of element positions from 
each of its parents. 

PBX proceeds in five steps. It first generates a list mapping each element to its
indexes in the parents. Consider $p_1 = [2, 5, 1, 4, 3, 0]$ and $p_2 = [5, 4, 3, 2, 1, 0]$.
The list of index mappings would be $[0 \rightarrow (5, 5), 1 \rightarrow (2, 4), 
2 \rightarrow (0, 3), 3 \rightarrow (4, 2), 4 \rightarrow (3, 1), 5 \rightarrow (1, 0)]$.
Mapping $1 \rightarrow (2, 4)$ means that element 1 is at index 2 in $p_1$ and index
4 in $p_2$. Step 2 randomizes the order of the elements in this list, e.g.,
$[3 \rightarrow (4, 2), 5 \rightarrow (1, 0), 0 \rightarrow (5, 5), 2 \rightarrow (0, 3), 
1 \rightarrow (2, 4), 4 \rightarrow (3, 1)]$. Also at this stage, PBX chooses a random subset of
elements (each element chosen with probability 0.5), and swaps the order of the indexes of those
elements. For this example, elements 5 and 1 are chosen. Mapping $5 \rightarrow (1, 0)$ becomes
$5 \rightarrow (0, 1)$, and $1 \rightarrow (2, 4)$ becomes $1 \rightarrow (4, 2)$. This results
in $[3 \rightarrow (4, 2), 5 \rightarrow (0, 1), 0 \rightarrow (5, 5), 2 \rightarrow (0, 3), 
1 \rightarrow (4, 2), 4 \rightarrow (3, 1)]$. Step 3 begins populating the children by iterating
this list, and for each mapping $e \rightarrow (i_1,i_2)$ it attempts to put element $e$ at index 
$i_1$ in $c_1$ and index $i_2$ in $c_2$, skipping any if the index is occupied. For this example,
we'd have: $c_1 = [5, x, x, 4, 3, 0]$ and $c_2 = [x, 5, 3, 2, x, 0]$. Step 4 makes a second pass
trying the index from the other parent to obtain: $c_1 = [5, x, 1, 4, 3, 0]$ and 
$c_2 = [x, 5, 3, 2, 1, 0]$. One final pass places any remaining elements in open indexes: 
$c_1 = [5, 2, 1, 4, 3, 0]$ and $c_2 = [4, 5, 3, 2, 1, 0]$.

PBX is strongly position oriented, although less so than CX since PBX's last pass puts some 
elements at different indexes than either parent. But unlike CX, PBX children inherit equal 
numbers of element positions from parents on average. As a result, children also inherit 
approximately equal numbers of precedences from parents, but many edges are disrupted 
since inherited positions are not grouped together. The runtime of PBX is $O(n)$. 

\textbf{Heuristic-Guided Crossover Operators:} We are primarily focusing on
crossover operators that are problem-independent, and which do not require
any knowledge of the optimization problem at hand. However, there are also 
some powerful problem-dependent crossover operators that utilize a heuristic
for the problem. We discuss a few of these here, although these are not
currently included in our open source library~\cite{cicirello2020joss}. One
of the more well known crossover operators of this type is Edge Assembly
Crossover (EAX)~\cite{Watson1998} for the TSP, which utilizes a TSP specific 
heuristic. Many have proposed variations and improvements to 
EAX~\cite{Sanches2017,Nagata2013,Nagata2006}, including adapting to other 
problems~\cite{Nagata2010,Nagata2007}. Another example of a crossover operator
guided by a heuristic is Heuristic Sequencing Crossover 
(HeurX)~\cite{cicirello2010flairs}, which was originally designed for
scheduling problems, and which requires a constructive heuristic for the
problem. EAX and its many variations are focused on problems where edges
are most important to solution fitness, whereas HeurX is primarily 
focused on precedences. There are other crossover operators that rely on
problem-dependent information~\cite{Freisleben1996}, and utilize local search while creating
children~\cite{Barecke2022}.

\textbf{Algorithmic Complexity:} The runtime (worst and average cases) of the crossover 
operators in this section, except the heuristic operators, is $O(n)$.

\section{\uppercase{Crossover Landscapes}}
\label{sec:xoverlandscapes}

\noindent In this section, we analyze the fitness landscape characteristics of
the crossover operators. We use the Permutation in a Haystack 
problem~\cite{cicirello2016evc} to define five artificial fitness landscapes, each 
isolating one of the permutation features listed earlier in Section~\ref{sec:assumptions}. 
In the Permutation in a Haystack, we must search for the permutation that 
minimizes distance to a target permutation. The optimal solution is obviously the 
target, much like in the OneMax problem the optimal solution is the bit string of 
all one bits. However, the Permutation in a Haystack provides a mechanism for 
isolating a permutation feature of interest in how we define distance. To focus 
on element positions, we use exact match distance~\cite{ronald1998}, the number of 
elements in different positions in the permutations. We define fitness landscapes for 
undirected and directed edges using cyclic edge distance~\cite{ronald1997} and 
cyclic r-type distance~\cite{Campos2005}, respectively. Cyclic edge distance is the 
number of edges in the first permutation, but not in the other, if the permutations 
represent undirected edges, including an edge from last element to first. Cyclic 
r-type distance is its directed edge counterpart. We use Kendall tau 
distance~\cite{kendall1938}, the minimum number of adjacent swaps to transform one 
permutation into the other, to define a fitness landscape for precedences. We use 
Lee distance~\cite{lee58} to define a landscape characterized by cyclic precedences. 
This last permutation feature, and corresponding distance function, were derived 
from a principal component analysis in our prior 
work~\cite{cicirello2019bict,cicirello2022mone}, and we are unaware of any real 
problem where this feature is important. 

For all landscapes, we use permutation length $n=100$, and generate 100 random
target permutations. We use an adaptive EA that evolves crossover and mutation 
rates to eliminate tuning these a priori. Each population member 
$(p_i, c_i, m_i, \sigma_i)$ consists of a permutation $p_i$, crossover rate 
$c_i$, mutation rate $m_i$, and a $\sigma_i$. In a generation, parent permutations 
$p_i$, $p_j$ cross with probability $c_i$ (which parent is $i$ is arbitrary), and 
each permutation mutates with probability $m_i$. We use swap mutation because it 
is simple and works reasonably well across a range of features. The $c_i$ and 
$m_i$ are initialized randomly in $[0.1, 1.0]$, and mutated with Gaussian mutation 
with standard deviation $\sigma_i$, constrained to $[0.1, 1.0]$. The $\sigma_i$ are 
initialized randomly in $[0.05, 0.15]$ and mutated with Gaussian mutation with 
standard deviation 0.01, constrained to $[0.01, 0.2]$. The population size is 100, 
and we use elitism with the single highest fitness solution surviving unaltered to 
the next generation. We use binary tournament selection. As a baseline, we use an 
adaptive mutation-only EA with swap mutation to examine whether the addition of a 
crossover operator improves performance. For UOBX, OX2, and UPPX, we use $u=0.5$ as 
suggested by their authors; and we use $u=0.33$ for UPMX for the same reason.

We use OpenJDK 17, Windows 10, an AMD A10-5700 3.4 GHz CPU, and 8GB RAM.
We use Chips-n-Salsa version 6.4.0, as released via the Maven Central repository, 
and not a development version to ensure reproducibility. The distance metrics
for the Permutation in a Haystack are from version 5.1.0
of JavaPermutationTools (JPT)~\cite{cicirello2018joss}. The source code 
of Chips-n-Salsa (https://github.com/cicirello/Chips-n-Salsa) and 
JPT (https://github.com/cicirello/JavaPermutationTools) is on GitHub; as is the 
source of the experiments, and the raw and processed data 
(https://github.com/cicirello/permutation-crossover-landscape-analysis).

\begin{figure}[t]
  \centering
   {\epsfig{file = 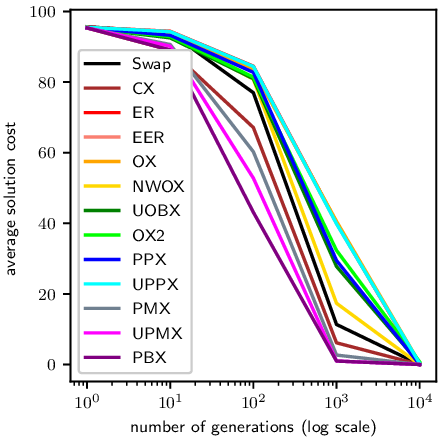, width = 2.95in}}
  \caption{Crossover comparison for element positions.}
  \label{fig:positions}
\end{figure}

\begin{figure}[t]
  \centering
   {\epsfig{file = 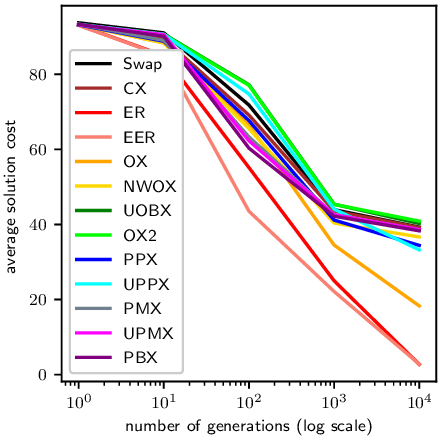, width = 2.95in}}
  \caption{Crossover comparison for undirected edges.}
  \label{fig:undirected}
\end{figure}

\begin{figure}[t]
  \centering
   {\epsfig{file = 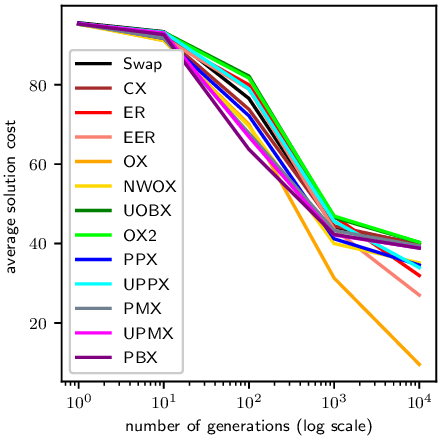, width = 2.95in}}
  \caption{Crossover comparison for directed edges.}
  \label{fig:directed}
\end{figure}

\begin{figure}[t]
  \centering
   {\epsfig{file = 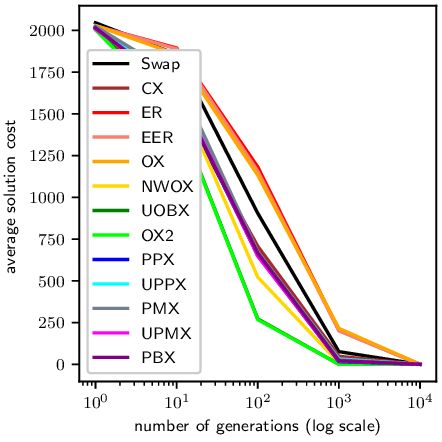, width = 2.95in}}
  \caption{Crossover comparison for precedences.}
  \label{fig:tau}
\end{figure}

\begin{figure}[t]
  \centering
   {\epsfig{file = 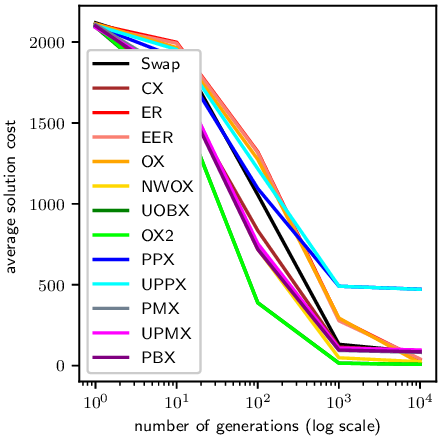, width = 2.95in}}
  \caption{Crossover comparison for cyclic precedences.}
  \label{fig:lee}
\end{figure}

Figures~\ref{fig:positions} to~\ref{fig:lee} show the results on the five landscapes.
The $x$-axis is number of generations at log scale, and the $y$-axis is solution cost
averaged over 100 runs. A black line shows the mutation-only baseline to make it
easy to see which crossover operators enable improvement over the mutation-only case.
Subtle differences among them are not particularly important, as we are only identifying
which crossover operators are worth considering for a problem based upon permutation 
features. Inspect the raw and summary data in GitHub for a fine-grained comparison.

\textbf{Optimizing element positions:} In Figure~\ref{fig:positions}, we see four crossover 
operators optimize element positions better than mutation alone: CX, PMX, UPMX, and PBX. 
Although UOBX and OX2 perform worse than baseline, they may also be relevant for positions 
if $u$ is carefully tuned as discussed in Section~\ref{sec:crossover}.

\textbf{Optimizing undirected edges:} Figure~\ref{fig:undirected} shows that only EER, ER, 
and OX optimize undirected edges better than the mutation-only baseline.

\textbf{Optimizing directed edges:} For directed edges (Figure~\ref{fig:directed}), only 
OX performs substantially better than the mutation-only baseline. For long runs, others 
(EER, ER, NWOX, PPX, UPPX) begin to outperform the baseline, but by much smaller margins 
than OX.

Although not seen here, with carefully tuned $u$, UOBX, OX2, and UPMX may be relevant 
for edges (undirected or directed) as discussed in Section~\ref{sec:crossover}.

\textbf{Optimizing precedences:} Several crossover operators optimize precedences
(Figure~\ref{fig:tau}) better than the mutation-only baseline, including: CX, NWOX, 
UOBX, OX2, PPX, UPPX, PMX, UPMX, and PBX.

\textbf{Optimizing cyclic precedences:} Figure~\ref{fig:lee} shows the cyclic
precedences results, which reveal many crossover operators superior to mutation 
alone: CX, NWOX, UOBX, OX2, PMX, UPMX, and PBX. Note that some of these strongly
overlap on the graph.

\begin{table*}[t]
\caption{Evolutionary operator characteristics: \checkmark means effective for feature, 
and ? means may be effective if carefully tuned.}\label{tab:summary}
\centering
\begin{tabular}{l|ccccc|c} \hline
         &           & Undirected & Directed &             & Cyclic      &  Limited \\
Operator & Positions & Edges      & Edges    & Precedences & Precedences & /Special  \\\hline
CX       & \checkmark &          &           & \checkmark & \checkmark & \\
ER       &           & \checkmark &        &             &             & \\
EER       &           & \checkmark &        &             &             & \\
OX       &           & \checkmark & \checkmark &             &          & \\
NWOX       &           &          &           & \checkmark & \checkmark & \\
UOBX       & ?       &   ?    &   ?   & \checkmark & \checkmark & \\
OX2       & ?       &   ?    &   ?   & \checkmark & \checkmark & \\
PPX       &           &           &          & \checkmark &          & \\
UPPX       &        &    ?   &   ?   & \checkmark &          & \\
PMX       & \checkmark &         &           & \checkmark & \checkmark & \\
UPMX       & \checkmark & ? & ? & \checkmark & \checkmark & \\
PBX      & \checkmark &          &          & \checkmark & \checkmark & \\
EAX      &           & \checkmark & \checkmark &          &        &  \\
HeurX    &           &              &              & \checkmark &     & \\
\hline
Swap     & \checkmark & \checkmark & \checkmark & \checkmark & \checkmark & \\
Adjacent Swap & & & & & & \checkmark \\
Insertion & & \checkmark & \checkmark & \checkmark & \checkmark & \\
Reversal & & \checkmark & & & & \\
$2$-change & & \checkmark & & & & \\
3opt & & \checkmark & & & & \\
Block-Move & & \checkmark & \checkmark & & & \\
Block-Swap & & \checkmark & \checkmark & & & \\
$\mathit{Cycle}(\mathit{kmax})$ & \checkmark & ? & ? & ? & ? & \\
$\mathit{Cycle}(\alpha)$ & \checkmark & ? & ? & ? & ? & \\
Scramble & & & & \checkmark & & \\
Uniform Scramble & ? & ? & ? & ? & ? & \\
Rotation & & & & & & \checkmark \\
\hline
\end{tabular}
\end{table*}

\section{\uppercase{Conclusions}}
\label{sec:conclusion}

\noindent Table~\ref{tab:summary} maps the evolutionary operators to the 
permutation features they effectively optimize. The top and bottom parts of 
Table~\ref{tab:summary} focus on crossover and mutation operators, 
respectively. The crossover operator to feature mapping in the top of 
Table~\ref{tab:summary} is as derived in Section~\ref{sec:xoverlandscapes}. 
The mutation operator to feature mapping in the bottom of Table~\ref{tab:summary} 
is partially derived from our prior research~\cite{cicirello2022mone}. In some 
cases (indicated with a question mark), applicability of an evolutionary operator 
for a feature may require tuning a control parameter, such as $u$ for uniform 
scramble or the various uniform crossover operators. 

One objective of this paper is to serve as a sort of catalog of evolutionary 
operators relevant to evolving permutations. Another objective is to offer 
insights into which operators are worth considering for a problem based upon 
the characteristics of the problem. If a fitness function for a problem is 
heavily influenced by a specific permutation feature, the insights from this 
paper can assist in narrowing the available operators to those most likely 
effective. If a fitness function is influenced by a combination of features, 
then an operator that balances its behavior with respect to that combination 
of features may be a good choice.

In future work, we plan to dive deeper into the behavior of the operators 
whose strengths are less clear. The crossover operators UOBX, OX2, UPPX, and 
UPMX include a tunable parameter, as do the mutation operators uniform scramble, 
$\mathit{Cycle}(\mathit{kmax})$, and $\mathit{Cycle}(\alpha)$. Although our 
empirical analyses identified problem features for which these operators are 
well suited, we also hypothesized that it may be possible to tune their control
parameters for favorable performance on other permutation features. Future work 
will explore more definitively answering whether or not such tuning can extend 
the applicability of these operators to more features.

\bibliographystyle{apalike}
{\small
\bibliography{ecta2023}}

\end{document}